%% file: 1997-Fomicheva.tex
\documentclass[11pt,a4paper]{article}
\usepackage{times,latexsym}
\usepackage{url}
\usepackage[T1]{fontenc}
\usepackage{amsmath}
\usepackage{multirow}
\usepackage{graphicx}
\usepackage{amsmath, bm}
\usepackage{float}
\usepackage{booktabs}
\usepackage{amsfonts}
\usepackage{subcaption}
\newcommand{\CorrHat}[1]{\expandafter\hat#1}
\newcommand{\CorrOverline}[1]{\expandafter\overline#1}

\usepackage[acceptedWithA]{tacl2018v2}

\usepackage{xspace,mfirstuc,tabulary}

\newif\ifdraft 
\draftfalse
\newcommand{\revised}[1]{\ifdraft \textcolor{blue}{#1} \else #1 \fi}

\renewcommand{\vec}[1]{\mathbf{#1}}

\newif\iftaclinstructions
\taclinstructionsfalse % AUTHORS: do NOT set this to true
\iftaclinstructions
\renewcommand{\confidential}{}
\renewcommand{\anonsubtext}{(No author info supplied here, for consistency with
TACL-submission anonymization requirements)}
\newcommand{\instr}
\fi

\iftaclpubformat % this "if" is set by the choice of options

\else

\fi

\title{Unsupervised Quality Estimation for Neural Machine Translation}
\author{\\
 Marina Fomicheva,\textsuperscript{1} Shuo Sun,\textsuperscript{2} Lisa Yankovskaya,\textsuperscript{3} Fr\'{e}d\'{e}ric Blain,\textsuperscript{1} Francisco Guzm\'an,\textsuperscript{5}\\ Mark Fishel,\textsuperscript{3} Nikolaos Aletras,\textsuperscript{1} Vishrav Chaudhary,\textsuperscript{5} Lucia Specia\textsuperscript{1,4}\\
 \textsuperscript{1}University of Sheffield 
 \textsuperscript{2}Johns Hopkins University 
 \textsuperscript{3}University of Tartu \\
 \textsuperscript{4}Imperial College London
 \textsuperscript{5}Facebook AI \\
  \texttt{\textsuperscript{1}\{m.fomicheva,f.blain,n.aletras,l.specia\}@sheffield.ac.uk}\\
  \texttt{\textsuperscript{2}\{ssun32\}@jhu.edu} 
  \texttt{\textsuperscript{3}\{lisa.yankovskaya,fishel\}@ut.ee}\\
  \texttt{\textsuperscript{5}\{fguzman,vishrav\}@fb.com}\\
}

\date{}

\begin{document}
\maketitle
\begin{abstract}

% Quality Estimation (QE) is an important component in making Machine Translation (MT) useful in real-world applications. Existing approaches require large amounts of expert annotated data, computation and time for training. As an alternative, we devise an unsupervised approach to QE where no training or access to additional resources besides the MT system itself is required. By exploiting methods for uncertainty quantification, we achieve substantial gains in correlation with human judgments of quality, \revised{rivalling state-of-the-art supervised QE models.}
% significantly outperforming state-of-the-art in supervised QE. % system.
% To evaluate our approach we collect the first dataset that enables further work on unsupervised QE.

Quality Estimation (QE) is an important component in making Machine Translation (MT) useful in real-world applications, as it is aimed to inform the user on the quality of the MT output at test time. Existing approaches require large amounts of expert annotated data, computation and time for training. As an alternative, we devise an unsupervised approach to QE where no training or access to additional resources besides the MT system itself is required. Different from most of the current work that treats the MT system as a black box, we explore useful information that can be extracted from the MT system as a by-product of translation. By employing methods for uncertainty quantification, we achieve very good correlation with human judgments of quality, rivalling state-of-the-art supervised QE models. To evaluate our approach we collect the first dataset that enables work on both black-box and glass-box approaches to QE.
\end{abstract}

\input{introduction_FINAL}
\input{related_work}

\input{unsupervised_qe}

\input{data}
\input{experiments_FINAL}

\input{discussion_FINAL}
\input{conclusions}

\section*{Acknowledgments}
Marina Fomicheva, Lisa Yankovskaya, Fr\'{e}d\'{e}ric Blain, Mark Fishel, Nikolaos Aletras and Lucia Specia were supported by funding from the Bergamot project (EU H2020 Grant No. 825303).

\bibliography{references}
\bibliographystyle{acl_natbib}

\end{document}

%% file: introduction_FINAL.tex
\section{Introduction}

With the advent of neural models, Machine Translation (MT) systems have made substantial progress, reportedly achieving near-human %translation
quality for high-resource language pairs \cite{hassan2018achieving,barrault2019findings}. %MT is used in an increasing number of applications, with more users relying on it for practical needs. 
However, translation quality %provided by state-of-the-art (SOTA) MT systems
is not consistent across language pairs, domains and datasets. This is problematic for low-resource scenarios, where there is not enough training data and translation quality significantly lags behind. Additionally, neural MT (NMT) systems can be deceptive to the end user as they can generate fluent translations that differ 
% which are different
in meaning from the original \cite{bentivogli2016neural,castilho2017neural}. Thus, it is crucial to have a feedback mechanism to inform users about the trustworthiness of a given MT output. % having a mechanism to predict and inform the user on what extent a given MT output can be trusted or relied upon becomes crucial.

Quality estimation (QE) aims to predict the quality of the output provided by an MT system at test time when no gold-standard human translation is available. State-of-the-art (SOTA) QE models %are neural-based, and % learning algorithms, 
require large amounts of parallel data for pre-training and in-domain translations annotated with quality labels for training \cite{kim-lee-na:2017:WMT,fonseca2019findings}. However, such large collections of data are only available for a small set of languages in limited domains. 

Current work on QE typically treats the MT system as a black box. In this paper we propose an alternative glass-box approach to QE which allows us to address the task as an {\bf unsupervised problem}. % In this paper, we propose to treat QE as an {\bf unsupervised problem}. 
We posit that encoder-decoder NMT models \cite{sutskever2014sequence,bahdanau2014neural,vaswani2017attention} offer a rich source of information for directly estimating translation quality: (a) the \emph{output probability distribution} from the NMT system (i.e. the probabilities obtained by applying the softmax function over the entire vocabulary of the target language); and (b) the \emph{attention mechanism} used during decoding. Our assumption is that %we first assume that 
the more confident the decoder is, the higher the quality of the translation. 

While sequence-level probabilities of the top MT hypothesis have been used for confidence estimation in statistical MT \cite{specia2013quest,blatz2004confidence}, the output probabilities from deep Neural Networks (NNs) are generally not well calibrated, i.e. not representative of the true likelihood of the predictions \cite{nguyen2015posterior,guo2017calibration,lakshminarayanan2017simple}. Moreover, softmax output probabilities tend to be \emph{overconfident} and can assign a large probability mass to predictions that are far away from the training data \cite{gal2016dropout}. To overcome such deficiencies, we propose ways to exploit output distributions beyond the %probability of the 
top-1 prediction by exploring \emph{uncertainty quantification} methods for better probability estimates \cite{gal2016dropout,lakshminarayanan2017simple}.  
In our experiments, we account for different factors that can affect the reliability of model probability estimates in NNs, such as model architecture, training and search %\cite{guo2017calibration,ott2018analyzing}.
\cite{guo2017calibration}.

In addition, we study attention mechanism as another source of information on NMT quality. Attention can be interpreted as a soft alignment, providing an indication of the strength of relationship between source and target words \cite{bahdanau2014neural}. While this interpretation is straightforward for NMT based on Recurrent Neural Networks (RNN) \cite{rikters2017confidence}, its application to current SOTA Transformer models with multi-head attention \cite{vaswani2017attention} is challenging. We analyze to what extent meaningful information on translation quality can be extracted from multi-head attention.

To evaluate our approach in challenging settings, we collect a new dataset for QE with 6 language pairs representing NMT training in high, medium, and low-resource scenarios. To reduce the chance of overfitting to particular domains, our dataset is constructed from Wikipedia documents. We annotate 10K segments per language pair. By contrast to the vast majority of work on QE that uses semi-automatic metrics based on post-editing distance as gold standard, % thus focusing on the specific quality aspects that are relevant for post-editing purposes, we evaluate how well QE methods are able to predict overall MT quality. To that end,
we perform quality labelling based on the Direct Assessment (DA) methodology \cite{da_paper}, which has been widely used for popular MT evaluation campaigns in the recent years. At the same time, the collected data % we collect
differs from the existing datasets annotated with DA judgments for the well known WMT Metrics task\footnote{\url{http://www.statmt.org/wmt19/metrics-task.html}} in two important ways: we provide enough data to train supervised QE models and access to the NMT systems used to generate the translations, thus allowing for further exploration of the glass-box unsupervised approach to QE for NMT introduced in this paper.

Our {\bf main contributions} can be summarised as follows: (i) A new, large-scale dataset for sentence-level\footnote{While the paper covers QE at sentence level, the extension of our unsupervised metrics to word-level QE would be straightforward and we leave it for future work.} QE annotated with DA rather than post-editing metrics (\S \ref{subsec:data}); (ii) A set of unsupervised quality indicators that can be produced as a by-product of NMT decoding and a thorough evaluation of how they correlate with human judgments of translation quality (\S \ref{sec:unsupervised_quality_estimation} and \S \ref{sec:experiments}); (iii)  The first attempt at analysing the attention distribution for the purposes of unsupervised QE in Transformer models (\S \ref{sec:unsupervised_quality_estimation} and \S \ref{sec:experiments}); (iv) The analysis on how model confidence relates to translation quality for different NMT systems (\S \ref{sec:discussion}). Our experiments show that unsupervised QE indicators obtained from well-calibrated NMT model probabilities \revised{rival strong supervised SOTA models in terms of correlation with human judgments.}

% we assess how well QE methods can predict general MT quality

% perform equally well or better than SOTA supervised QE approaches.

% Our results show that unsupervised QE indicators obtained from well-calibrated NMT model probabilities outperform state-of-the-art supervised QE approaches when the quality distribution of the NMT translations is not too skewed.

%% file: related_work.tex
\section{Related Work}\label{related_work}

% Niko's general comments for related work:

% - Subsec 1 {Supervised QE}: Introduce previous work on QE using supervised approaches. Introduce what HTER is etc.. Limitations: you need training data, problems with low resource languages, might be noisy/expensive, you need a separate model to your NMT model.

% - Subsec 2 (Unsupervised QE): Introduce efforts for unsupervised QE: 

% \paragraph{Model Confidence}: Source and target language model probabilities have been successfully used as indicators to predict translation QE following the assumption that a more confident model produces better translations than a less confident one but not for NMT systems. Second paragraph: Just what you have in current 2.3
    
% \paragraph{Attention-based approaches:} Just what you have under 2.2
    
% Finish with something like: To the best of our knowledge this is the first attempt to unsupervised QE using info from pretrained NMT systems.

\paragraph{QE}
% By difference to the well-known automatic evaluation metrics that compare the MT output to a human translation to assess its quality such as BLEU \cite{papineni2002bleu} QE systems aim to predict the quality of MT at test time, when no human reference is available for comparison. SOTA models for QE are supervised and require a significant amount of in-domain labelled data %, in addition to unlabelled parallel data \cite{kim-lee-na:2017:WMT,openkiwi}; they do not explore any internal information from the NMT model. This information has been used for QE of statistical MT systems in feature-based models (the so-called {\em glass-box features})  \cite{blatz2004confidence,specia2013quest}. %Namely, source and target language model probabilities have been successfully used as indicators of model confidence to predict translation quality. NMT system information has not, however, thus far been explored for QE and no previous work addresses unsupervised QE for NMT.

QE is typically addressed as a supervised machine learning task where the goal is to predict MT quality in the absence of % gold-standard
reference translation. Traditional feature-based approaches relied on manually designed features, % either
extracted from the MT system (\emph{glass-box} features) or obtained from the source and translated sentences, as well as external resources, such as monolingual or parallel corpora (\emph{black-box} features) \cite{specia2009estimating}.

Currently the best performing approaches to QE  employ NNs to learn useful representations for source and target sentences \cite{kim2017predictor,wang-etal-2018-alibaba,kepler2019unbabel}. A notable example is the Predictor-Estimator (PredEst) model~\cite{kim2017predictor}, which consists of an encoder-decoder RNN ({\it predictor}) trained on parallel data for a word prediction task and a unidirectional RNN ({\it estimator}) that produces quality estimates leveraging the context representations generated by the predictor. % \newcite{kepler2019unbabel} reused this architecture alongside very large-scale pretrained representations from BERT~\cite{devlin2018bert} and XLM~\cite{lample2019cross} to establish new SOTA results ~\cite{fonseca2019findings}.\ls{I wonder if we need the last sentence about XLM/BERT as we don't use this version in this paper. We could ciite them along with the predictor-estimator citation above. Let's chat on slack}
Despite achieving strong performances, neural-based approaches % to QE
are resource-heavy and require a significant amount of in-domain labelled data for training. They do not use any internal information from the MT system.

% The first noticeable architecture is the Predictor-Estimator (PredEst) model~\cite{kim2017predictor}, an encoder-decoder RNN ({\it predictor}), which predicts words along with context representation from an attention mechanism. These representations are used as input for a unidirectional RNN ({\it estimator}) to produce quality estimates.
% \newcite{kepler2019unbabel} reused this architecture alongside very large-scale pretrained representations from BERT~\cite{devlin2018bert} and XLM~\cite{lample2019cross}, as well as ensembling techniques, to establish new SOTA at the Fourth Conference on Machine Translation (WMT'19) QE Shared Task~\cite{fonseca2019findings}.}
% \revised{Despite achieving strong performances, neural-based approaches to QE are resource-heavy and require a significant amount of in-domain labelled data for training. They do not use any internal information from the MT system.}

Existing work on glass-box QE is limited to features extracted from statistical MT, such as language model (LM) probabilities or number of hypotheses in the n-best list
\cite{blatz2004confidence,specia2013quest}. The few approaches for unsupervised QE are also inspired by the work on statistical MT and perform significantly worse than supervised approaches \cite{popovic2012morpheme,moreau2012quality,etchegoyhen2018supervised}. For example, \citet{etchegoyhen2018supervised} use %utilise
lexical translation probabilities from word alignment models and LM probabilities. %Both lexical translation probabilities and n-gram language model are trained on external data. 
%Besides traditional feature combination, 
Their unsupervised approach average these features to produce the final score. However, it is largely outperformed by the %SOTA
neural-based supervised QE systems \cite{specia2018findings}.

% One exception to this trend is the work that uses the entropy of attention weight of RNN-based MT systems \cite{bahdanau2014neural} as an indicator of translation quality ~\cite{rikters2017confidence,yankovskaya-2018:WMT}.
The only work that explore internal information from neural models as an indicator of translation quality rely on the entropy of attention weights in RNN-based NMT systems ~\cite{rikters2017confidence,yankovskaya-2018:WMT}. However, attention-based indicators perform competitively only when combined with other QE features in a supervised framework. Furthermore, this approach is not directly applicable to the SOTA Transformer model that employs multi-head attention mechanism. Recent work on attention interpretability showed that attention weights in Transformer networks might not be readily interpretable \cite{vashishth2019attention,vig2019analyzing}. \citet{voita2019analyzing} show that different attention heads of Transformer 
% play different roles: positional, syntactic and rare words; also, some heads are more important than others.
have different functions and some of them are more important than others. This makes it challenging to extract information from attention weights in Transformer (see \S \ref{sec:experiments}). %Sections \ref{subsec:attention} and \ref{subsubsec:correlation_analysis}. }

To the best of our knowledge, our work is the first on glass-box unsupervised QE for NMT that performs competitively with respect to the SOTA supervised systems.

\paragraph{QE Datasets}
The performance of QE systems has been typically assessed using the semi-automatic HTER (Human-mediated Translation Edit Rate) \cite{snover2006study} metric as gold standard. However, the reliability of this metric for assessing the performance of QE systems has been shown to be questionable \cite{graham2016all}. The current practice in MT evaluation is the so called Direct Assessment (DA) of MT quality \cite{da_paper}, where raters evaluate the MT on a continuous 1-100 scale. This method has been shown to improve the reproducibility of manual evaluation and to provide a more reliable gold standard for automatic evaluation metrics \cite{graham2015accurate}.

DA methodology is currently used for manual evaluation of MT quality at the WMT %News and other
translation tasks, as well as for assessing the performance of reference-based automatic MT evaluation metrics at the WMT Metrics Task \cite{bojar-etal-2016-results,bojar-etal-2017-results,ma-etal-2018-results,ma-etal-2019-results}. Existing datasets with sentence-level DA judgments from the WMT Metrics Task could in principle be used for benchmarking QE systems. However, they contain only a few hundred segments % 560 segments \ls{560 is one specific year. Should we not say they contain only a few hundred segments?}
per language pair and thus hardly allow for training supervised systems, as illustrated by the weak correlation results for QE on DA judgments based on the Metrics Task data recently reported by \citet{fonseca2019findings}. Furthermore, for each language pair the data contains translations from a number of MT systems often using different architectures, and these MT systems are not readily available, making it impossible for experiments on glass-box QE. Finally, the judgments are either crowd-sourced or collected from task participants and not professional translators, which may hinder the reliability of the labels. We collect a new dataset for QE that addresses these limitations (\S \ref{subsec:data}).

\paragraph{Uncertainty quantification}
Uncertainty quantification in NNs is typically addressed using a Bayesian framework where the point estimates of their weights are replaced with probability distributions \cite{mackay1992bayesian,graves2011practical,welling2011bayesian,tran2019bayesian}. Various approximations have been developed to avoid high training costs of Bayesian NNs, such as Monte Carlo Dropout \cite{gal2016dropout} or model ensembling \cite{lakshminarayanan2017simple}. %Simple post-processing methods that transform model probabilities after generation, such as temperature scaling \cite{platt1999probabilistic}, are another alternative to improve model confidence estimates \cite{guo2017calibration}.
The performance of uncertainty quantification methods is commonly evaluated by measuring calibration, i.e. the relation between predictive probabilities and the empirical frequencies of the predicted labels, or by assessing generalization of uncertainty under domain shift (see \S \ref{sec:discussion}).

Only a few studies analyzed calibration in NMT and they came to contradictory conclusions. \citet{kumar2019calibration} measure calibration error by comparing model probabilities and the percentage of times NMT output matches reference translation, and conclude that NMT probabilities are poorly calibrated. However, the calibration error metrics they use are designed for binary classification tasks and cannot be easily transferred to NMT %, which involves structured prediction with very large output space
\cite{kuleshov2015calibrated}. \citet{ott2019fairseq} analyze uncertainty in NMT by comparing predictive probability distributions with the empirical distribution observed in human translation data. They conclude that NMT models are well calibrated. However, this approach is limited by the fact that there are many possible correct translations for a given sentence and only one human translation is available in practice. Although the goal of this paper is to devise an unsupervised solution for the QE task, the analysis presented here provides new insights into calibration in NMT. Different from existing work, we study the relation between model probabilities and human judgments of translation correctness.

% Only a few studies have analyzed calibration in NMT but they came to contradictory conclusions. \citet{kumar2019calibration} measures calibration error by comparing model probabilities and the percentage of times NMT output matches reference translation and conclude that NMT probabilities are poorly calibrated. \citet{ott2019fairseq} analyse uncertainty in NMT by comparing predictive probability distributions with the empirical distribution observed in human translation data. They conclude that NMT models are well calibrated. Although the goal of this paper is to devise an unsupervised solution for the QE task, the analysis presented here provides new insights into calibration in NMT. Different from existing work

Uncertainty quantification methods have been successfully applied to various practical tasks, e.g. neural semantic parsing \cite{dong2018confidence},  hate speech classification \cite{miok2019prediction}, or back-translation for NMT \cite{wang2019improving}. \citet{wang2019improving}, which is the closest to our work, explore a small set of uncertainty-based metrics to minimise the weight of erroneous synthetic sentence pairs for back translation in NMT. However, %we note that
improved NMT training with weighted synthetic data does not necessarily imply better prediction of MT quality. In fact, metrics that \citet{wang2019improving} report to perform the best for back-translation do not perform well for QE (see \S \ref{subsec:uncertainty}).

%% file: unsupervised_qe.tex
\section{Unsupervised QE for NMT}
\label{sec:unsupervised_quality_estimation}

% In this paper we assume 
%In this Section we introduce the basics of NMT, as well as some necessary terminology and notation that will be used throughout the paper.
%Most work on 
% NMT models based on the sequence-to-sequence approach with encoder and decoder networks \cite{sutskever2014sequence}.

We assume %that the NMT model is based on
a sequence-to-sequence NMT architecture consisting of encoder-decoder networks using attention \cite{bahdanau2014neural}. The encoder maps the input sequence $\vec{x}={x_1,...,x_I}$ into a sequence of hidden states, %$\vec{z}={z_1,...,z_I}$,
which is summarized into a single vector using attention mechanism \cite{bahdanau2014neural,vaswani2017attention}. Given this representation %$\vec{z}$
the decoder generates an output sequence $\vec{y}={y_1,...,y_T}$ of length $T$. The probability of generating $\vec{y}$ is factorized as:
\begin{equation}
p(\vec{y}|\vec{x},\theta) = \prod_{t=1}^{T} p(y_t|\vec{y}_{<t},\vec{x},\theta)
\nonumber
\end{equation}
where $\theta$ represents model parameters. The decoder produces the probability distribution $p(y_t|\vec{y}_{<t},\vec{x},\theta)$ over the system vocabulary at each time step using the \emph{softmax function}. The model is trained to minimize cross-entropy loss. %:
%\begin{equation}
%\mathcal{L}=-\sum_{t=1}^T \log p(y_t|\vec{y}_{<t},\vec{x},\theta)
%\end{equation}
%
% While several neural network architectures have been proposed for NMT,  %\cite{bahdanau2014neural,luong2015effective,gehring2017convolutional,vaswani2017attention}, 
% we focus on SOTA Transformers \cite{vaswani2017attention} as architecture.   
%While the metrics defined below can be extracted from any encoder-decoder NMT model with attention, 
We use SOTA Transformers \cite{vaswani2017attention} for the encoder and decoder in our experiments.

%. Throughout Sections \ref{sec:unsupervised_quality_estimation} and \ref{sec:experiments} we present unsupervised quality indicators that can be extracted from any attention-based encoder-decoder NMT model, whereas 
%In Section  
%\ref{subsec:results_uncertainty_models} we show some comparison against other NMT variants. % to study whether model architecture and training affects the correlation of output probabilities with perceived translation quality.

% Niko's comments:

% 1. We need to disentagle unsupervised QE metrics from specific NMT model architectures. You can start with a paragraph saying: We assume that an NMT model consisting of an encoder, decoder, encoder attention, decoder attention. That could be a separate section before this one actually, a shorter version of Section 3.

% 2. We use information from the above components or their outputs. We need to make it clear.

% 3. Use 3 subsections: (1) Confidence/Uncertainty based metrics; (2) Attention-based metrics; (3) Similarity between source and translation metrics.

% 4. Make sure that we highlight metric names (e.g. \paragraph{} or \nonident {\bf ...}) at the beginning of each paragraph.

% 5. Make sure that metric names and abbreviations are well defined and consistent through out the paper. This is very important for presenting the results.

In what follows,
we propose unsupervised quality indicators based on: (i) output probability distribution obtained either from a standard deterministic NMT (\S \ref{subsec:probability}) or (ii) using uncertainty quantification (\S \ref{subsec:uncertainty}), and (iii) attention weights (\S \ref{subsec:attention}). %We note that the measures are defined for sentence-level but can be easily extended to word-level QE.

% Below we introduce various measurements that can be extracted from the predicted probability distributions generated during standard decoding for each translated word.

%In this paper we aim to devise an unsupervised approach to QE. 
% In what follows we explore strategies to improve model calibration at inference time \todo{improve model calibration or improve uncertainty estimation? do we really improve it or just use methods to estimate it?} based on output probability distribution (Section \ref{subsec:probability}) and attention weights (Section \ref{subsec:attention}). %, whereas the next section is focused on how model architecture and training settings affect the accuracy of model predictive probabilities.

\input{softmax_proba}

\input{uncertainty_FINAL}
\input{attention}

%% file: softmax_proba.tex
% \subsection{Model Confidence}
\subsection{Exploiting the Softmax Distribution}
\label{subsec:probability}

% \todo{this is confusing as we are talking about confidence and calibration at the same time. should we say here we assume the model is well calibrated, then in the next section we talk about how to try to make sure it is?}
% We argue that one useful source of information for QE of NMT models is model confidence. 

% The same metrics can be obtained using a determinisitc NMT model or a probabilistic NMT model by applying MC-dropout

% In a well calibrated model probabilities associated with a prediction are already a good indicator of the likelihood of its accuracy \cite{niculescu2005predicting,naeini2015obtaining,nguyen2015posterior,guo2017calibration}. Based on this premise, we define a simple QE measure for NMT using sequence-level translation probability, normalized by the number of output target tokens $T$:

We start by defining a simple QE measure based on sequence-level translation probability normalized by length: % the number of output tokens:
\begin{equation}
\mathrm{TP} = \frac{1}{T} \sum_{t=1}^{T} \log p(y_t|\vec{y}_{<t},\vec{x},\theta)
\nonumber
\end{equation}

However, 1-best probability estimates from the softmax output distribution may tend towards overconfidence, which would result in high probability for unreliable MT outputs. We propose two metrics that exploit output probability distribution beyond the average of top-1 predictions. First, we compute the entropy of softmax output distribution over target vocabulary of size ${V}$ at each decoding step and take an average to obtain a sentence-level measure:
\begin{equation}
\mathrm{Softmax\mbox{-}Ent} = -\frac{1}{T}\sum_{t=1}^{T}\sum_{v=1}^{V} p(y_t^v) \log p(y_t^v)
\nonumber
\end{equation}
where $p(y_t)$ represents the conditional distribution $p(y_{t}|\vec{x},\vec{y}_{<t},\theta)$.

If most of the probability mass is concentrated on a few vocabulary words, the generated target word is likely to be correct. By contrast, if softmax probabilities approach a uniform distribution picking any word from the vocabulary is equally likely and the quality of the resulting translation is expected to be low.

Second, we hypothesize that the dispersion of probabilities of individual words might provide useful information that is inevitably lost when taking an average. Consider, as an illustration, that the sequences of word probabilities [0.1, 0.9] and [0.5, 0.5] have the same mean, but might indicate very different behaviour of the NMT system, and consequently, different output quality. To formalize this intuition we compute the standard deviation of word-level log-probabilities.
\begin{align}
\mathrm{Sent\mbox{-}Std} = \sqrt{\mathbb{E}[\mathrm{P}^2]-(\mathbb{E}[\mathrm{P}])^2}
\nonumber
\end{align}

\noindent where $\mathrm{P}=p(y_1),...,p(y_T)$ represents word-level log-probabilities for a given sentence.

% \begin{align}
% \mathrm{Sent\mbox{-}Std} = \sqrt{\mathrm{Sent\mbox{-}Var}}
% \nonumber
% \end{align}

% \begin{align}
% \mathrm{Sent\mbox{-}Var} = \frac{1}{T} \sum_{t=1}^{T} [\log p(y_t)]^2 -  \mathbb{E}[\log p(y_t)]^2
% \nonumber
% \end{align}

% \begin{align}
% \mathrm{Sent\mbox{-}Std} = \sqrt{\frac{1}{1-T}\sum_{t=1}^{T}(\log p(y_t)-\mathrm{TP})^2}
% \nonumber
% \end{align}

% \revised{where $p(y_t)$ represents the conditional distribution $p(y_t|\vec{x},\vec{y}_{<t},\theta)$. }
% \vc{noob question: Shouldn't the first term be}
% \math{E[log p(y_t)^2]}

% \begin{equation}
% \begin{split}
% \mathrm{Sent\mbox{-}Var} = \frac{1}{T} \sum_{t=1}^{T} \log p(y_{t})^2 - \mathbb{E}[\log {p(y_t)}]^2
% \end{split}
% \end{equation}

%% file: uncertainty_FINAL.tex
\subsection{Quantifying Uncertainty}\label{subsec:uncertainty}

% Above we assumed the NMT probabilities are well calibrated and therefore can be a good indicator of MT output quality. This might not be the case, as recent studies suggest that modern neural networks may be poorly calibrated \cite{nguyen2015posterior,guo2017calibration,lakshminarayanan2017simple}. Superior results for unsupervised QE can be achieved by improving uncertainty quantification. We explore a widely used approach to approximate Bayesian inference proposed by \citet{gal2016dropout}. We also look at a simple non-Bayesian method for calibrating predictive probabilities called temperature scaling \cite{platt1999probabilistic,guo2017calibration}. 

It has been argued in recent work that deep neural networks do not properly represent model uncertainty \cite{gal2016dropout,lakshminarayanan2017simple}. Uncertainty quantification in deep learning typically relies on the Bayesian formalism \cite{mackay1992bayesian,graves2011practical,welling2011bayesian,gal2016dropout,tran2019bayesian}. Bayesian NNs learn a posterior distribution over parameters that quantifies \emph{model} or \emph{epistemic} uncertainty, i.e. our lack of knowledge as to which model generated the training data.\footnote{A distinction is typically made between epistemic and aleatoric uncertainty, where the latter captures the noise inherent to the observations \cite{kendall2017uncertainties}. We leave modelling aleatoric uncertainty in NMT for future work.} 
Bayesian NNs usually come with prohibitive computational costs and various approximations have been developed to alleviate this. % issue. %The one we explore in this paper is
In this paper we explore the
{\bf Monte Carlo (MC) dropout} \cite{gal2016dropout}. 
%To avoid changing the NMT training procedure, instead of using Bayesian networks we explore an approximation to Bayesian inference called {\bf Monte Carlo (MC) dropout} \citet{gal2016dropout}.

%\paragraph{Monte Carlo Dropout}

%A standard way for reasoning about uncertainty in machine learning is to use Bayesian methods.
%Bayesian neural networks usually come with prohibitive computational costs and various approximations have been developed to alleviate this issue. One such approach developed by \citet{gal2016dropout} is called Monte Carlo (MC) dropout. 
Dropout is a method introduced by \citet{srivastava2014dropout} to reduce overfitting when training neural models. It consists in randomly masking neurons to zero based on a Bernoulli distribution. \citet{gal2016dropout} use dropout at test time before every weight layer. They perform several forward passes through the network and collect posterior probabilities generated by the model with parameters perturbed by dropout. Mean and variance of the resulting distribution can then be used to represent model uncertainty. %They show that this is mathematically equivalent to an approximation to the probabilistic deep Gaussian process.

We propose two flavours of MC dropout-based measures for unsupervised QE. \textbf{First}, we compute the expectation and variance for the set of sentence-level probability estimates obtained by running $N$ stochastic forward passes through the MT model with model parameters $\hat{\theta}$ perturbed by dropout: %, as we have observed that this results in an improved performance in our experiments.}}%\paco{we need the correct $_n$ index}
% \sshuo{Minor comment: Why do we need take the absolute values over the Ns? I thought N is always positive.}

\begin{equation}
\mathrm{D\mbox{-}TP} = \frac{1}{N} \sum_{n=1}^{N} \mathrm{TP}_{\hat{\theta}^{n}}
\nonumber
\end{equation}
\begin{equation}
\mathrm{D\mbox{-}Var} = \mathbb{E}[\mathrm{TP}^2_{\hat{\theta}}]-(\mathbb{E}[\mathrm{TP}_{\hat{\theta}}])^2
\nonumber
\end{equation}

% where $\widehat{TP}$ is defined as $\log p(\vec{y}^{(n)}|\vec{x},\hat{\theta})$.

% \begin{equation}
% \mathrm{D\mbox{-}TP} = \frac{1}{N} \sum_{n=1}^{N} \log p(\vec{y}^{(n)}|\vec{x},\hat{\theta})
% \nonumber
% \end{equation}
% \begin{equation}
% \mathrm{D\mbox{-}Var} = \frac{1}{N} \sum_{n=1}^{N} [\log p(\vec{y}^{(n)}|\vec{x},\hat{\theta})]^2 -  \mathrm{D\mbox{-}TP}^2
% \nonumber
% \end{equation}

\noindent where $\mathrm{TP}$ is sentence-level probability as defined in \S \ref{subsec:probability}. We also look at a combination of the two:
\begin{equation}
\mathrm{D\mbox{-}Combo} = \Big(1 - \frac{\mathrm{D\mbox{-}TP}}{\mathrm{D\mbox{-}Var}}\Big)
\nonumber
\end{equation}

% \noindent where $\mathrm{TP}$ is the sentence-level probability normalized by sentence-length, as defined in \S \ref{subsec:probability}.

% where $\hat{\theta}$ represents model parameters perturbed by dropout.

We note that these metrics have also been used by \citet{wang2019improving}, but with the purpose of minimising the effect of low quality outputs on NMT training with back translations.

\textbf{Second}, we measure lexical variation between the MT outputs generated for the same source segment when running inference with dropout. We posit that differences between likely MT hypotheses may also capture uncertainty and potential ambiguity and complexity of the original sentence. We compute an average similarity score ($sim$) between the set $\mathbb{H}$ of translation hypotheses:
%
% \begin{equation}
% \mathrm{D\mbox{-}Ling\mbox{-}Sim} = C \sum_{h}^{H} \mathrm{METEOR}(h^i,h^j), i \neq j
% \end{equation}
%
\begin{equation}
\mathrm{D\mbox{-}Lex\mbox{-}Sim} = \frac{1}{C}\sum_{i=1}^{|\mathbb{H}|} \sum_{j=1}^{|\mathbb{H}|} sim(h_i,h_j)
\nonumber
\end{equation}
% \noindent where $h_i, h_j \in H$, $i \ne j$ and $C=2^{-1}|H|(|H|-1)$ is the number of pairwise comparisons for $|H|$ hypotheses. $sim$ corresponds to the similarity function. We use Meteor \cite{denkowski2014meteor} to compute similarity scores.%, but any other similarity metric can be used for this purpose. 
\noindent where $h_i, h_j \in \mathbb{H}$, $i \ne j$ and $C=2^{-1}|\mathbb{H}|(|\mathbb{H}|-1)$ is the number of pairwise comparisons for $|\mathbb{H}|$ hypotheses. We use Meteor \cite{denkowski2014meteor} to compute similarity scores.

%% file: attention.tex
\subsection{Attention}
\label{subsec:attention}

% Attention weights show the strength of connection between source and target tokens and this strength might indicate translation quality \cite{rikters2017confidence}. One way to estimate the strength of connection or how much attention target tokens pay to source tokens is to compute the entropy of the attention distribution:

Attention weights represent the strength of connection between source and target tokens, which may be indicative of translation quality \cite{rikters2017confidence}. One way to measure it is to compute the entropy of the attention distribution:

\begin{equation}
\mathrm{Att\mbox{-}Ent} = -\frac{1}{I}\sum_{i=1}^I\sum_{j=1}^J \alpha_{ji} \log\alpha_{ji}  
\nonumber
\end{equation}
where $\alpha$ represents attention weights, $I$ is the number of target tokens and $J$ is the number of source tokens. %\vc{Add information about alpha}

This mechanism can be applied to any NMT model with encoder-decoder attention. We focus on attention in Transformer models, as it is currently the most widely used NMT architecture. Transformers rely on various types of attention, multiple attention heads and multiple encoder and decoder layers. Encoder-decoder attention weights are computed for each head (H) and for each layer (L) of the decoder, as a result we get $[H \times L]$ matrices with attention weights. It is not clear which combination would give the best results for QE. %We propose the following ways to summarize the information from different heads and layers.
To summarize the information from different heads and layers, we propose to compute the entropy scores for each possible head/layer combination and then choose the minimum value or compute the average:
\begin{equation}
    \mathrm{AW\mbox{:}Ent\mbox{-}Min} =  min_{\{hl\}}(\mathrm{Att\mbox{-}Ent_{hl})} 
\nonumber
\end{equation}
\begin{equation}
    \mathrm{AW\mbox{:}Ent\mbox{-}Avg} = \frac{1}{H\times L}\sum_{h=1}^{H}\sum_{l=1}^{L}\mathrm{Att\mbox{-}Ent_{hl}}
\nonumber
\end{equation}

%% file: data.tex
\section{Multilingual Dataset for QE}
\label{subsec:data}

The quality of NMT translations is strongly affected by the amount of training data. To study our unsupervised QE indicators under different conditions, we collected data for 6 language pairs that includes high-, medium-, and low-resource conditions. To add diversity, we varied the directions into and out-of English, when permitted by the availability of expert annotators into non-English languages. Thus our dataset is composed by the high-resource English--German (En-De) and  English--Chinese (En-Zh) pairs; by the medium-resource Romanian--English (Ro-En) and Estonian--English (Et-En) pairs; and by the low-resource Sinhala--English (Si-En) and Nepali--English (Ne-En) pairs. The dataset contains sentences extracted from Wikipedia and the MT outputs manually annotated for quality.

%To conduct our experiments, we collected a new dataset for QE. %Below, we describe the process followed to sample and clean the data, as well as to generate translations and annotate them.

\paragraph{Document and sentence sampling}
We follow the sampling process outlined in FLORES \cite{guzman-etal-2019-flores}.
First, we sampled documents from Wikipedia for English, Estonian, Romanian, Sinhala and Nepali. Second, we selected the top $100$ documents containing the largest number of sentences that are: (i) in the intended source language according to a language-id classifier\footnote{\url{https://fasttext.cc}} and (ii) have the length between 50 and 150 characters. In addition, we filtered out sentences that have been released as part of recent Wikipedia parallel corpora \cite{schwenk2019wikimatrix} ensuring that our dataset is not part of parallel data commonly used for NMT training.

For every language, we randomly selected 10K sentences from the sampled documents and then translated them into English using the MT models described below. For German and Chinese we selected 20K sentences from the top 100 documents in English Wikipedia. To ensure sufficient representation of high- and low-quality translations for high-resource language pairs, we selected the sentences with minimal lexical overlap with respect to the NMT training data. %\mf{We should may present DA distribution in this section and say that for English-German even in these conditions it is highly skewed as most of the translations are high quality (see my plot and discussion in Figure \ref{fig:distribution_across_languages}.}

\paragraph{NMT systems}
%\textbf{NMT systems.}

For medium- and high-resource language pairs we trained the MT models based on the standard Transformer architecture \cite{vaswani2017attention} and followed the implementation details described in \citet{ott2018scaling}. We used publicly available MT datasets such as Paracrawl \cite{espla-etal-2019-paracrawl} and Europarl \cite{koehn2005europarl}. Si-En and Ne-En MT systems were trained based on Big-Transformer architecture as defined in \cite{vaswani2017attention}. For the low-resource language pairs, the models were trained following the FLORES semi-supervised setting \cite{guzman-etal-2019-flores}\footnote{\url{https://bit.ly/36YaBlU}} which involves two iterations of backtranslation using the source and the target monolingual data. Table \ref{tab:annotations} specifies the amount of data used for training.

% Si-En and Ne-En MT systems use the semi-supervised setting\footnote{https://bit.ly/36YaBlU} for training
% We trained MT models for all language pairs based on the standard Transformer architecture \cite{vaswani2017attention} and followed the implementation details described in \citet{ott2018scaling}.
%as mentioned in \citet{guzman-etal-2019-flores}. For Nepali-English and Sinhala-English, 
% we trained MT models with the dataset used for training the baseline systems in \citet{guzman-etal-2019-flores}.
% For medium- and high-resource language pairs we trained MT models with publicly available MT datasets such as Paracrawl \cite{espla-etal-2019-paracrawl} and Europarl \cite{koehn2005europarl}. Table \ref{tab:annotations} specifies the amount of data used for training.
% For Nepali-English and Sinhala-English, we trained MT models with the dataset used for training  the baseline systems in \citet{guzman-etal-2019-flores}, which contains around 564K and 647K parallel sentences for Nepali-English and Sinhala-English respectively. For Romanian-English and Estonian-English we trained MT models with publicly available MT datasets such as Paracrawl \cite{espla-etal-2019-paracrawl} and Europarl \cite{koehn2005europarl}. The training datasets were composed of more than 20M parallel sentences for English-German and English-Chinese models, and around 4M and 900K parallel sentences for Estonian-English and Romanian-English models, respectively.

\paragraph{DA judgments} 
%\textbf{DA judgments.}
We followed the FLORES setup \cite{guzman-etal-2019-flores}, which presents a form of DA \cite{graham-EtAl:2013:LAW7-ID}. The annotators are asked to rate each sentence from 0-100 according to the perceived translation quality. Specifically, the 0-10 range represents an incorrect translation; 11-29, a translation with few correct keywords, but the overall meaning is different from the source; 30-50,  a translation with major mistakes; 51-69, a translation which is understandable and conveys the overall meaning of the source but contains typos or grammatical errors; 70-90, a translation that closely preserves the semantics of the source sentence; and 90-100, a perfect translation.

% \revised{Each sentence was evaluated by up to 6 professional translators from two language service providers (3 translators each). To improve annotation consistency, within each language service provider any evaluation in which the range of scores among the raters was above 30 points was rejected, and an additional rater was requested to replace the most diverging translation rating until convergence was achieved.}

Each segment was evaluated independently by three professional translators from a single language service provider. To improve annotation consistency, any evaluation in which the range of scores among the raters was above 30 points was rejected, and an additional rater was requested to replace the most diverging translation rating until convergence was achieved. To further increase the reliability of the test and development partitions of the dataset, we requested an additional set of three annotations from a different group of annotators (i.e. from another language service provider) following the same annotation protocol, thus resulting in 
a total of six annotations per segment. %for the development and test data.}

Raw human scores were converted into \emph{z-scores}, i.e. standardized according to each individual annotator's overall mean and standard deviation. The scores collected for each segment were averaged to obtain the final score. Such setting allows for the fact that annotators may genuinely disagree on some aspects of quality.%\ls{do we need the last sentence? I'm not sure it adds much}}

In Table~\ref{tab:annotations} we show a summary of the statistics from human annotations. Besides the NMT training corpus size and the distribution of the DA scores for each language pair, we report mean and standard deviation of the average differences between the scores assigned by different annotators to each segment, as an indicator of annotation consistency. First, we observe that, as expected, the amount of training data per language pair correlates with the average quality of an NMT system. Second, we note that the distribution of human scores changes substantially across language pairs. In particular, we see very little variability in quality for En-De, which makes QE for this language pair especially challenging (see \S \ref{sec:experiments}). Finally, as shown in the right-most columns, annotation consistency is similar across language pairs and comparable to existing work that follows DA methodology for data collection. For example, \citet{graham-EtAl:2013:LAW7-ID} reports an average difference of 25 across annotators' scores.

% \mf{We should add: agreement results, screenshot of the interface, histograms of the DA scores} %We observe that the amount of training data per language pair correlates with the average quality of a system.

\paragraph{Data splits}
%\textbf{Data splits.}
To enable comparison between supervised and unsupervised approaches to QE, we split the data into 7K training partition, 1K development set, and two testsets of 1K sentences each. One of these testsets is used for the experiments in this paper, the other is kept blind for future work.  

\paragraph{Additional data}
%\textbf{Additional data for Estonian-English.}
To support our discussion of the effect of NMT training on the correlation between predictive probabilities and perceived translation quality presented in \S \ref{sec:discussion}, we trained various alternative NMT system variants, translated and annotated 400 original Estonian sentences from our test set with each system variant.

% \revised{To facilitate further comparison with existing work, we provide two additional types of data. On the one hand, for Et-En we collected human reference translation for the 1K test partition to allow for a comparison with reference-based automatic evaluation metrics. On the other hand, in order to reproduce current common practices in QE we collected post-editing data for training, development and test partitions for En-De and En-Zh.\footnote{The selection of the language pairs in this case is motivated by the availability of expert annotation.} \mf{We need to decide whether we add reference-based metrics and HTER to the results.}}

\input{tables/mlqe_dataset}
% \begin{table}[htb]
% \centering
%\footnotesize
% \small
% \begin{tabular}{l  c c c  c c c} 
% \toprule
%& & \multicolumn{5}{c}{\textbf{DA} }\\\cmidrule{3-6}
% \textbf{Lang} & \textbf{c.size} & & avg & p25 & median & p75 \\
% \midrule
% \multicolumn{7}{l}{\bf High-resource} \\
% En-De & 23.7M & & 87.1 & 84.6 & 89.6 & 93.0\\
% En-Zh & 22.6M & & 68.9 & 61.0 & 74.6 & 82.0\\
% \midrule
% \multicolumn{7}{l}{\bf Mid-resource} \\
% Ro-En & 3.9M & & 68.4 & 48.6 & 76.0 & 92.6\\
% Et-En & 880K & & 67.2 & 44.3 & 75.6 & 90.0\\
% \midrule
% \multicolumn{7}{l}{\bf Low-resource} \\
% Si-En & 647K & & 51.0 & 21.6 & 51.8 & 80.0\\ 
% Ne-En & 564K & & 38.7 & 23.0 & 34.8 & 51.6\\ 
% \bottomrule
% \end{tabular}
%  \caption{\label{tab:annotations} Multilingual QE dataset: size of the training corpus and summary statistics for the DA scores.}
 %corpus size of the NMT system used to generate the hypotheses, along with summary statistics for the DA scores: the average (avg), 25th percentile, median and 75th percentile of the raw DA scores.}
% \end{table}

The data, the NMT models and the DA judgments are available at \url{https://github.com/facebookresearch/mlqe}.  %Thus, this dataset will provide access to reliable human judgments of MT outputs and NMT models that generated the translations. On the one hand, this information is required for the development of unsupervised QE and, to the best of our knowledge, is not readily available as part of any other QE dataset. On the other hand, it contains a sufficient amount of data with human annotation allowing for training supervised QE systems. 
%We hope it will become an invaluable resource  for the development and evaluation of both supervised and unsupervised approaches to QE.

% \todo{we should introduce the fact that we have more data for et-en and the subset of questions we can answer for the other languages for which we do not have the model variants}

%% file: tables/mlqe_dataset.tex
\begin{table*}[htb]
\centering
\small
\begin{tabular}{l l l c c c c c c} 
\toprule
& & & \multicolumn{4}{c}{{\bf scores}} & \multicolumn{2}{c}{\bf diff}\\
\cmidrule(r){4-7}\cmidrule(l){8-9}
& \textbf{Pair} & {\bf size} & { avg} & { p25} & { median} & { p75} & {avg} & { std} \\
\midrule
%\multicolumn{6}{l}
\multirow{ 2}{*}{\bf High-resource} &
{\bf En-De} & 23.7M & 84.8 & 80.7 & 88.7 & 92.7 & 13.7 & 8.2 \\
& {\bf En-Zh} & 22.6M & 67.0 & 58.7 & 70.7 & 79.0 & 12.1 & 6.4 \\
\midrule
\multirow{ 2}{*}{\bf Mid-resource} &
{\bf Ro-En} & 3.9M & 68.8 & 50.1 & 76.0 & 92.3 & 10.7 & 6.7 \\
& {\bf Et-En} & 880K & 64.4 & 40.5 & 72.0 & 89.3 & 13.8 & 9.4 \\
\midrule
\multirow{ 2}{*}{\bf Low-resource}&
{\bf Si-En} & 647K & 51.4 & 26.0 & 51.3 & 77.7 & 13.4 & 8.7 \\ 
& {\bf Ne-En} & 564K & 37.7 & 23.3 & 33.7 & 49.0 & 11.5 & 5.9 \\ 
\bottomrule
\end{tabular}
% \caption{\label{tab:annotations} Multilingual QE dataset: size of the training corpus and summary statistics for the DA scores.}
\caption{\label{tab:annotations} \revised{Multilingual QE dataset: size of the NMT training corpus (size) and summary statistics for the raw DA scores (average, 25th percentile, median and 75th percentile). As an indicator of annotators' consistency, the last two columns show the mean (avg) and standard deviation (std) of the absolute differences (diff) between the scores assigned by different annotators to the same segment.}}
%corpus size of the NMT system used to generate the hypotheses, along with summary statistics for the DA scores: the average (avg), 25th percentile, median and 75th percentile of the raw DA scores.}
\end{table*}

%% file: experiments_FINAL.tex
\section{Experiments and Results}
\label{sec:experiments}
Below we analyze how our unsupervised QE indicators correlate with human judgments.

\subsection{Settings}
\label{subsec:settings}

% \paragraph{Data} We use the dataset and NMT models described in \S \ref{subsec:data}. To make the results comparable with supervised QE, we use the test (1K) partition of the dataset for evaluation, whereas train (7K) and dev (1K) partitions are used to train the supervised QE systems.

\paragraph{Benchmark supervised QE systems}
%\mf{We need to add here information on other supervised models that we use: BERT-BiRNN, BERT-Estimator, NuQE?}\todo{LS: maybe we add this pred-est description to the related work instead}

We compare the performance of the proposed unsupervised QE indicators against the best performing supervised approaches with available open-source implementation, namely the Predictor-Estimator (PredEst) architecture~\cite{kim2017predictor} provided by OpenKiwi toolkit~\cite{openkiwi}, and an improved version of the BiRNN model provided by DeepQuest toolkit~\cite{iveetal-deepquest2018}, which we refer to as BERT-BiRNN~\cite{blain_tlqe}.

\textbf{PredEst.} %this model is an encoder-decoder RNN ({\it predictor}), which predicts words along with context representation from an attention mechanism. These representations are used as input for a unidirectional RNN ({\it estimator}) that produces quality estimates. 
We trained PredEst models (see \S \ref{related_work}) using the same parameters as in the default configurations provided by~\citet{openkiwi}. Predictor models were trained for 6 epochs on the same training and development data as the NMT systems, while the Estimator models were trained for 10 epochs on the training and development sets of our dataset (see \S \ref{subsec:data}). Unlike~\citet{openkiwi}, the Estimator was not trained using multi-task learning, as our dataset currently does not contain any word-level annotation. %For each of the two models, 
We use the model corresponding to the best epoch as identified by the metric of reference on the development set: perplexity for the Predictor and Pearson correlation for the Estimator.

\textbf{BERT-BiRNN.} This model, similarly to the recent SOTA QE systems \cite{kepler2019unbabel}, %(see \S \ref{related_work}),
uses a large scale pre-trained BERT model to obtain token-level representations which are then fed into two independent bidirectional RNNs to encode both the source sentence and its translation independently. The two resulting sentence representations are then concatenated as a weighted sum of their word vectors, using an attention mechanism. The final sentence-level representation is then fed to a sigmoid layer to produce the sentence-level quality estimates. %\ls{Fred to add how: what's the output layer (that does the regression)?}
% Compared to PredEst, the BERT-BiRNN model %is a lightweight architecture that does not use any extra parallel data for training, yet it achieves comparable results, as shown in performances (anonymous). \ls{not sure we should say that since we do use BERT and Unbabel uses openkiwi without Pred but with BERT. I would remove it.} ~\cite{okabe2020mqe}.
% The models were trained on the training and development partitions of the dataset (see \S \ref{subsec:data}). \todo{we don't need this last sentence, it's sort of a given}%, using token-level representations for a pretrained multilingual cased base BERT model.
During training, BERT was fine-tuned by unfreezing the weights of the last four layers along with the embedding layer. We used early stopping based on Pearson correlation on the development set, with a patience of 5.

\input{tables/summary_table_all_languages_FINAL}

\paragraph{Unsupervised QE}
For the dropout-based indicators (see \S \ref{subsec:uncertainty}), we use dropout rate of 0.3, the same as for training the NMT models (see \S \ref{subsec:data}). We perform $N=30$ inference passes to obtain the posterior probability distribution. $N$ was chosen following the experiments in related work \cite{dong2018confidence,wang2019improving}. However, we note that increasing $N$ beyond 10 results in very small improvements on the development set. The implementation of stochastic decoding with MC dropout is available as part of the fairseq toolkit \cite{ott2019fairseq} at \url{https://github.com/pytorch/fairseq}. % \url{https://github.com/fairseq/tree/TACL2020_mc_dropout/examples/unsupervised_quality_estimation}.}

\subsection{Correlation with Human Judgments}

Table \ref{tab:uncertainty_indicators} shows Pearson correlation with DA for our unsupervised QE indicators %described in Section \ref{sec:unsupervised_quality_estimation},
and for the supervised QE systems. 
%and the reference-based metrics presented above. 
Unsupervised QE indicators are grouped as follows: \textbf{Group I} corresponds to the measurements obtained with standard decoding (\S \ref{subsec:probability}); \textbf{Group II} contains indicators computed using MC dropout (\S \ref{subsec:uncertainty}); 
% \footnote{We use dropout rate of 0.3 and apply it before each weight layer. We perform $N=30$ inference passes to obtain the posterior probability distribution.}
and \textbf{Group III} contains the results for attention-based indicators (\S \ref{subsec:attention}). \textbf{Group IV} corresponds to the supervised QE models presented in \S \ref{subsec:settings}. We use the Hotelling-Williams test to compute significance of the difference between dependent correlations \cite{Williams:1959} with p-value $< 0.05$. For each language pair, results that are not significantly outperformed by any method are marked in bold; results that are not significantly outperformed by any other method from the same group are underlined.%\vc{What is our definition of significantly outperformed? In the first column both 0.513 and 0.473 are bold, whereas in the second column only 0.600 is bold (not 0.558)}

\begin{figure*}[ht]
\centering
\begin{subfigure}[t]{0.5\textwidth}
    \includegraphics[width=.9\textwidth]{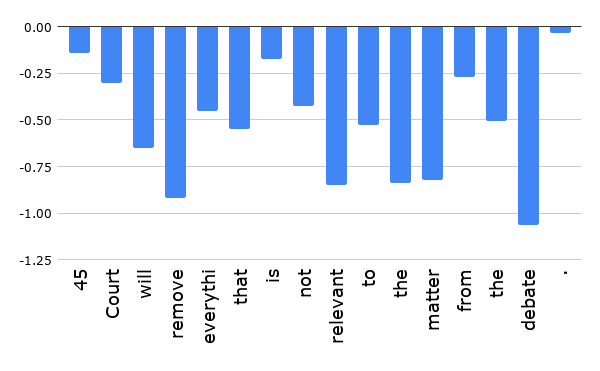}
\end{subfigure}%
\begin{subfigure}[t]{0.5\textwidth}
    \includegraphics[width=.9\textwidth]{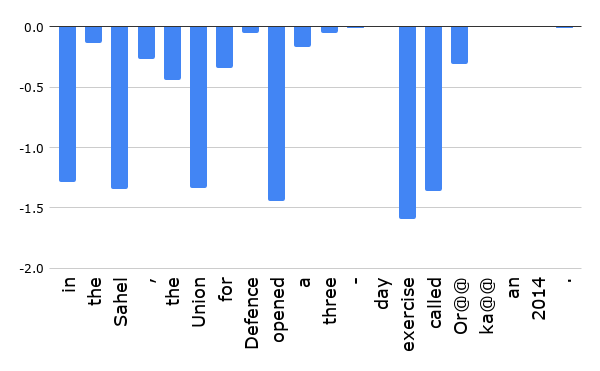}
\end{subfigure}
% \caption{High-quality (left) and low-quality (right) Et-En translations with very similar overall sentence-level probability (-0.50 and -0.48 for the first and second sentence, respectively) but very different patterns in the sequence of token-level probabilities.}
\caption{Token-level probabilities of high-quality (left) and low-quality (right) Et-En translations.}
\label{fig:token_probabilities}
\end{figure*}

\input{tables/example_variation_FINAL.tex}

We observe that the simplest measure that can be extracted from NMT, sequence-level probability (TP), already performs competitively, in particular for the medium-resource language pairs. TP is consistently outperformed by D-TP, indicating that NMT output probabilities are not well calibrated. This confirms our hypothesis that estimating model uncertainty improves correlation with perceived translation quality. Furthermore, our approach performs competitively with strong supervised QE models. Dropout-based indicators significantly outperform PredEst and rival BERT-BiRNN for four language pairs\footnote{We note that PredEst models are systematically and significantly outperformed by BERT-BiRNN. This is not surprising, as large-scale pretrained representations have been shown to boost model performance for QE \cite{kepler2019unbabel} and other natural language processing tasks \cite{devlin2018bert}.}. % \ls{maybe this is sufficient, instead of the note about the unbabel model in the introduction}}} 
These results position the proposed unsupervised QE methods as an attractive alternative to the supervised approach in the scenario where the NMT model used to generate the translations can be accessed. %, as they can be extracted as a by product of translation with no need to access any additional resources besides the NMT system itself. 

% Another general observation is that
For both unsupervised and supervised methods performance varies considerably across language pairs. The highest correlation is achieved for the medium-resource languages, whereas for high-resource language pairs it is drastically lower. The main reason for this difference is a lower variability in translation quality for high-resource language pairs. Figure \ref{fig:distribution_across_languages} shows %the distribution of DA scores and
scatter plots for Ro-En, which has the best correlation results, and En-De with the lowest correlation for all quality indicators. Ro-En has a substantial number of high-quality sentences, but the rest of the translations are uniformly distributed across the quality range. The distribution for En-De is highly skewed, as the vast majority of the translations are of high quality. In this case capturing meaningful variation appears to be more challenging, as the differences reflected by the DA may be more subtle than any of the QE methods is able to reveal. % \paco{It is interesting that here we're seeing the opposite effect as with the WMT datasets.}

The reason for a lower correlation for Sinhala and Nepalese is different. For unsupervised indicators it can be due to the difference in model capacity\footnote{Models for these languages were trained using Transformer-Big architecture from \citet{vaswani2017attention}.} 
%Transformer for the rest of the language pairs)\mf{Vishrav, could you please confirm this?}
and the amount of training data. On the one hand, increasing depth and width of the model may negatively affect calibration \cite{guo2017calibration}. On the other hand, due to the small amount of training data the model can overfit, resulting in inferior results both in terms of translation quality and correlation. It is noteworthy, however, that supervised QE system suffers a larger drop in performance than unsupervised indicators, as its predictor component requires large amounts of parallel data for training. We suggest, therefore, that unsupervised QE is more stable in low-resource scenarios than supervised approaches. 

We now look in more detail at the three groups of unsupervised measurements in Table \ref{tab:uncertainty_indicators}.

\paragraph{Group I} Average entropy of the softmax output ($\mathrm{Softmax\mbox{-}Ent}$) and dispersion of the values of token-level probabilities ($\mathrm{Sent\mbox{-}Std}$) achieve a significantly higher correlation than TP metric for four language pairs. 
% produce competitive results, with a substantial improvement over sequence-level translation probability for some of the language pairs.
$\mathrm{Softmax\mbox{-}Ent}$ captures uncertainty of the output probability distribution, which appears to be a more accurate reflection of the overall translation quality. $\mathrm{Sent\mbox{-}Std}$ captures a pattern in the sequence of token-level probabilities that helps detect low-quality translation illustrated in Figure \ref{fig:token_probabilities}. Figure \ref{fig:token_probabilities} shows two Et-En translations which have drastically different absolute DA scores of 62 and 1, but the difference in their sentence-level log-probability is negligible: -0.50 and -0.48 for the first and second translations, respectively. By contrast, the sequences of token-level probabilities are very different, as the second sentence has larger variation in the log-probabilities for adjacent words, with very high probabilities for high-frequency function words and low probabilities for content words.

% They have the absolute DA scores of 62 and 1, respectively, the first sentence being clearly superior in quality. But the difference in their sentence-level log-probability is negligible: -0.50 and -0.48 for the first and second translations, respectively. By contrast, the sequence of token-level probabilities for these two sentences looks very differently, as the second sentence has larger variation in the log-probabilities for adjacent words, with very high probabilities for high-frequency function words and low probabilities for content words.

%\todo{LS: this is another candidate to go if we need space as I'm not sure it adds much to the paper, even though it's super interesting. we may be able to explain without plots} %We suggest that further analysis of the patterns of token-level probabilities in translated sequences can lead to even better correlation results.

\paragraph{Group II} The best results are achieved by the $\mathrm{D\mbox{-}Lex\mbox{-}Sim}$ and $\mathrm{D\mbox{-}TP}$ metrics. % that measures the expectation over posterior probabilities obtained using MC dropout. %This confirms our hypothesis that a better representation of model uncertainty leads to a higher correlation with human judgements of MT quality. 
Interestingly, $\mathrm{D\mbox{-}Var}$ has a much lower correlation, since by only capturing variance it ignores the actual probability estimate assigned by the model to the given output.\footnote{This is in contrast with the work by \citet{wang2019improving} where $\mathrm{D\mbox{-}Var}$ appears to be one of the best performing metric for NMT training with back-translation demonstrating an essential difference between this task and QE.}

Table \ref{tab:linguistic_variation} provides an illustration of how model uncertainty captured by MC dropout reflects the quality of MT output. The first example contains a low quality translation, with a high variability in MT hypotheses obtained with MC dropout. By contrast, MC dropout hypotheses for the second high-quality example are very similar and, in fact, constitute valid linguistic paraphrases of each other. This fact is directly exploited by the $\mathrm{D\mbox{-}Lex\mbox{-}Sim}$ metric that measures the variability between MT hypotheses generated with perturbed model parameters and performs on pair with $\mathrm{D\mbox{-}TP}$. Besides capturing model uncertainty, $\mathrm{D\mbox{-}Lex\mbox{-}Sim}$ reflects the potential complexity of the source segments, as the number of different possible translations of the sentences is an indicator of their inherent ambiguity.\footnote{Note that $\mathrm{D\mbox{-}Lex\mbox{-}Sim}$ involves generating $N$ additional translation hypotheses, whereas the $\mathrm{D\mbox{-}TP}$ only requires re-scoring an existing translation output and is thus less expensive in terms of time.} %It should be noted, however, that this is a less practical method, since generating additional translations is more costly than re-scoring an already generated MT output multiple times.

% A question that arises for the practical application of $\mathrm{D\mbox{-}TP}$ is how the results would be affected by the number of stochastic forward passes. Figure \ref{fig:forward_passes} shows Pearson correlation with human judgments for our $\mathrm{D\mbox{-}TP}$ metric as a function of the number of stochastic forward passes of MC dropout for Romanian-English and Estonian-English, where this metric brings the largest gains over sequence-level probability. Based on the results in Figure \ref{fig:forward_passes}, we suggest using $N=20$ as increasing N beyond 20 does not give any improvement in correlation.

% \begin{figure}[ht]
%\centering
%     \includegraphics[width=0.45\textwidth]{figures/n_forward_passes.png}
% \caption{Pearson correlation with human judgments for our $\mathrm{D\mbox{-}TP}$ metric as a function of the number of stochastic forward passes of MC dropout for Ro-En and Et-En.}
% \label{fig:forward_passes}
% \end{figure}

\paragraph{Group III} While our attention-based metrics also achieve a sensible correlation with human judgments, it is considerably lower than the rest of the unsupervised indicators. Attention may not provide enough information to be used as a quality indicator of its own, since 
there is no direct mapping between words in different languages, and, therefore, high entropy in attention weights does not necessarily indicate low translation quality. We leave experiments with combined attention and probability-based measures to future work.

% On the other hand, 
The use of multi-head attention with multiple layers in Transformer may also negatively affect the results. As shown by \citet{voita2019analyzing}, different attention heads are responsible for different functions. Therefore, combining the information coming from different heads and layers in a simple way may not be an optimal solution. To test whether this is the case, we computed attention entropy and its correlation with DA for all possible combinations of heads and layers. As shown in Table \ref{tab:uncertainty_indicators}, the best head/layer combination ($\mathrm{AW\mbox{:}\text{best head/layer}}$) indeed significantly outperforms other attention-based measurements for all language pairs suggesting that this method should be preferred over simple averaging. Using the best head/layer combination for QE is limited by the fact that it requires validation on a dataset annotated with DA and thus is not fully unsupervised. This outcome opens an interesting direction for further experiments to automatically discover the best possible head/layer combination. 

% \revised{\paragraph{Group IV} Despite being trained on the same data as the NMT systems, PredEst models are systematically and significantly outperformed across all language-pairs by BERT-based QE models. This confirms that large scale pretrained representations are more powerful than representation trained with the task at hand.}

%\footnote{When computing entropy, we exclude end of sentence ``EOS'' mark, as it is not a real word, and renormalize the remaining attention weights.}

% Results are shown in Table \ref{tab:uncertainty_indicators} ($\mathrm{AW\mbox{:}\text{best head/layer}}$). Indeed, for all language pairs the best head/layer combination significantly outperforms other attention-based measurements suggesting that this method should be preferred over simple averaging.

\input{figures/distribution_across_languages.tex}

%% file: tables/summary_table_all_languages_FINAL.tex
\renewcommand{\arraystretch}{1.1}
\begin{table*}[t]
\begin{center}
\small
% \footnotesize
\begin{tabular}{l l  c c c c c c} 
%\hline
\toprule
& & \multicolumn{2}{c}{\bf Low-resource} & \multicolumn{2}{c}{\bf Mid-resource} & \multicolumn{2}{c}{\bf High-resource}\\\cmidrule(r){3-4}\cmidrule(lr){5-6}\cmidrule(l){7-8}
 &{\bf Method} & Si-En & Ne-En & Et-En & Ro-En & En-De & En-Zh\\
\midrule
% \multicolumn{8}{c}{Unsupervised QE} \\
% \hline
\multirow{3}{*}{\bf I}& $\mathrm{TP}$ & 0.399 & 0.482 & \underline{0.486} & \underline{0.647} & 0.208 & 0.257 \\
 & $\mathrm{Softmax\mbox{-}Ent}$ (-) & \underline{0.457} & \underline{0.528} & 0.421 & 0.613 & 0.147 & 0.251 \\
& $\mathrm{Sent\mbox{-}Std}$ (-) & 0.418 & 0.472 & 0.471 & 0.595 & \textbf{\underline{0.264}} & \underline{0.301} \\
\midrule
\multirow{4}{*}{\bf II} & $\mathrm{D\mbox{-}TP}$ & 0.460 & 0.558 & \underline{\textbf{0.642}} & \underline{0.693} & \textbf{\underline{0.259}} & \underline{\textbf{0.321}} \\
& $\mathrm{D\mbox{-}Var}$ (-) & 0.307 & 0.299 & 0.356 & 0.332 & 0.164 & 0.232 \\
& $\mathrm{D\mbox{-}Combo}$ (-) & 0.286 & 0.418 & 0.475 & 0.383 & 0.189 & 0.225 \\
& $\mathrm{D\mbox{-}Lex\mbox{-}Sim}$ & \underline{\textbf{0.513}} & \underline{\textbf{0.600}} & 0.612 & 0.669 & 0.172 & \underline{\textbf{0.313}} \\
\midrule
\multirow{3}{*}{\bf III} & $\mathrm{AW\mbox{:}Ent\mbox{-}Min}$ (-) & 0.097 & 0.265 & 0.329 & \underline{0.524} & 0.000 & 0.067 \\
& $\mathrm{AW\mbox{:}Ent\mbox{-}Avg}$ (-) & 0.10 & 0.205 & 0.377 & 0.382 & 0.090 & 0.112 \\
& $\mathrm{AW\mbox{:}\text{best head/layer}}$ (-) & \underline{0.255} & \underline{0.381} & \underline{0.416} & \underline{0.636} & \underline{0.241} & \underline{0.168} \\
\midrule
\multirow{2}{*}{\bf IV} & $\mathrm{PredEst}$ & 0.374 & 0.386 & 0.477 & 0.685 & 0.145 & 0.190 \\
& $\mathrm{BERT\mbox{-}BiRNN}$ & \underline{\textbf{0.473}} & \underline{0.546} & \underline{\textbf{0.635}} & \underline{\textbf{0.763}} & \underline{\textbf{0.273}} & \underline{\textbf{0.371}} \\
\bottomrule
\end{tabular}
\end{center}
\caption{Pearson ($r$) correlation between unsupervised QE indicators and human DA judgments. Results that are not significantly outperformed by any method are marked in bold; results that are not significantly outperformed by any other method from the same group are underlined.} 
\label{tab:uncertainty_indicators}
\end{table*}

%% file: tables/example_variation_FINAL.tex
\begin{table*}[!ht]
\center
\small
\begin{tabular}{cll}
\toprule
\multirow{7}{*}{\rotatebox{90}{
Low Quality}}&{\bf Original} & Tanganjikast p{\"u}{\"u}takse niiluse ahvenat ja kapentat. \\
&{\bf Reference} & Nile perch and kapenta are fished from Lake Tanganyika. \\
&{\bf MT Output} & There is a silver thread and candle from Tanzeri. \\
%\hline
\cmidrule(l){3-3}
&\multirow{4}{*}{\bf Dropout} & There will be a silver thread and a penny from Tanzer. \\
& & There is an attempt at a silver greed and a carpenter from Tanzeri. \\
& & There will be a silver bullet and a candle from Tanzer. \\
& & The puzzle is being caught in the chicken's gavel and the coffin.\\
\midrule
\multirow{7}{*}{\rotatebox{90}{
High Quality}} & {\bf Original} & Siis aga v{\~o}ib tekkida seesmise ja v{\"a}lise vaate vahele l{\~o}he. \\
& {\bf Reference} & This could however lead to a split between the inner and outer view.  \\
& {\bf MT Output} & Then there may be a split between internal and external viewpoints. \\\cmidrule(l){3-3}
&\multirow{4}{*}{\bf Dropout} &  Then, however, there may be a split between internal and external viewpoints.  \\
& & Then, however, there may be a gap between internal and external viewpoints. \\
& & Then there may be a split between internal and external viewpoints. \\
& & Then there may be a split between internal and external viewpoints. \\
\bottomrule
\end{tabular}
\caption{Example of MC dropout for a low-quality (top) and a high-quality (bottom) MT outputs.}
\label{tab:linguistic_variation}
\end{table*}

%% file: figures/distribution_across_languages.tex
\begin{figure}
\centering
\begin{subfigure}[t]{0.5\textwidth}
    \centering
    \includegraphics[scale=0.3]{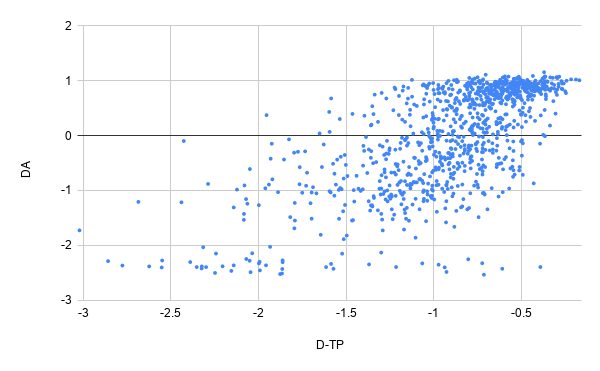}
\end{subfigure}
\begin{subfigure}[t]{0.5\textwidth}
    \centering
    \includegraphics[scale=0.3]{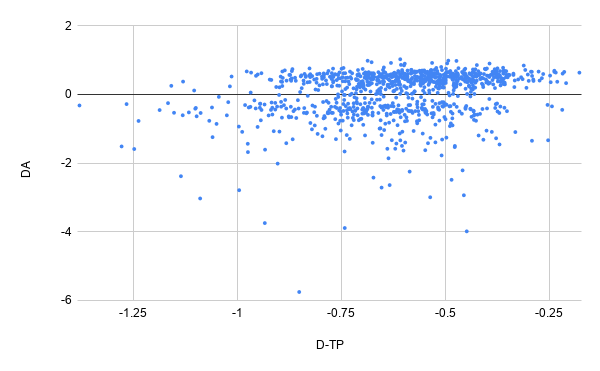}
\end{subfigure}
\caption{Scatter plots for the correlation between D-TP (x-axis) and standardized DA scores (y-axis) for Ro-En (top) and En-De (bottom).}
\label{fig:distribution_across_languages}
\end{figure}

%% file: discussion_FINAL.tex
% \section{Model Confidence across NMT Systems}
\section{Discussion}\label{sec:discussion}

% This Section provides a discussion of various questions related with experimental results presented above.

In the previous Section we studied the performance of our unsupervised quality indicators for different language pairs. In this Section we validate our results by looking at two additional factors: domain shift and underlying NMT system.

\subsection{Domain Shift}
% Here we assess whether the improvement in correlation with MC dropout is achieved due to a better quantification of uncertainty.
One way to evaluate how well a model represents uncertainty is to measure the difference in model confidence under domain shift \cite{hendrycks2016baseline,lakshminarayanan2017simple,ovadia2019can}. A well calibrated model should produce low confidence estimates when tested on data points that are far away from the training data. 

Overconfident predictions on out-of-domain sentences would undermine the benefits of unsupervised QE for NMT. This is particularly relevant given the current wide use of NMT for translating mixed domain data online. Therefore, we conduct a small experiment to compare model confidence on in-domain and out-of-domain data. We focus on the Et-En language pair. We use the test partition of the MT training dataset as our in-domain sample. To generate the out-of-domain sample, we sort our Wikipedia data (prior to sentence sampling stage in \S \ref{subsec:data}) by distance to the training data and select the top 500 segments with the largest distance score. To compute distance scores we follow the strategy %proposed by
of \citet{niehues2019modeling} that measures the test/training data distance based on the hidden states of NMT encoder. 

We compute model posterior probabilities for the translations of the in-domain and out-of-domain sample either obtained through standard decoding, or using MC dropout. TP obtains average values of -0.440 and -0.445 for in-domain and out-of-domain data respectively, whereas for D-TP these values are -0.592 and -0.685. The difference between in-domain and out-of-domain confidence estimates obtained by standard decoding is negligible. The difference between MC-dropout average probabilities for in-domain vs. out-of-domain samples was found to be statistically significant under Student's T-test, with p-value $< 0.01$. Thus, expectation over predictive probabilities with MC dropout indeed provides a better estimation of model uncertainty for NMT, and therefore can improve the robustness of unsupervised QE on out-of-domain data.

% D-TP stdev for out-domain: 0.26266426877094434
% D-TP stdev for in-domain: 0.20963651718668688

\subsection{NMT Calibration across NMT Systems}
\label{subsec:results_uncertainty_models}
Findings in the previous Section suggest that using model probabilities results in fairly high correlation with human judgments for various language pairs. In this Section we study how well these findings generalize to different NMT systems. The list of model variants that we explore is by no means exhaustive and was motivated by common practices in MT and by the factors that can negatively affect model calibration (number of training epochs) or help represent uncertainty (model ensembling). For this small-scale experiment we focus on Et-En. For each system variant we translated 400 sentences from the test partition of our dataset and collected the DA accordingly. As baseline, we use a standard Transformer model with beam search decoding. All system variants are trained using Fairseq implementation \cite{ott2019fairseq} for 30 epochs, with the best checkpoint chosen according to the validation loss.%\footnote{Unless explicitly indicated otherwise, all variants are based on Transformer model from \citet{vaswani2017attention}.}

\textbf{First}, we consider three system variants with differences in architecture or training:
%\begin{itemize}
%    \itemsep0em 
%    \item
RNN-based NMT \cite{bahdanau2014neural,luong2015effective},
%    \item
Mixture of Experts \cite[MoE,][]{he2018sequence,shen2019mixture,cho2019mixture} and
%    \item
model ensemble \cite{garmash2016ensemble}.
%\end{itemize}

\citet{shen2019mixture} use the \emph{MoE} framework to capture the inherent uncertainty of the MT task where the same input sentence can have multiple correct translations. A mixture model introduces a multinomial latent variable to control generation and produce a diverse set of MT hypotheses. In our experiment we use hard mixture model with uniform prior and 5 mixture components. To produce the translations we generate from a randomly chosen component with standard beam search. To obtain the probability estimates we average the probabilities from all mixture components.

Previous work has used \emph{model ensembling} as a strategy for representing model uncertainty \cite{lakshminarayanan2017simple,pearce2018uncertainty}.\footnote{Note that MC dropout discussed in \S \ref{subsec:uncertainty} can be interpreted as an ensemble model combination where the predictions are averaged over an ensemble of NNs \cite{lakshminarayanan2017simple}.} In NMT, ensembling has been used to improve translation quality. We train four Transformer models initialized with different random seeds. At decoding time predictive distributions from different models are combined by averaging.

\textbf{Second}, we consider two alternatives to beam search: diverse beam search \cite{vijayakumar2016diverse} and sampling. For sampling, we generate translations one token at a time by sampling from the model conditional distribution $p(y_j|\vec{y}_{<j},\vec{x},\theta)$, until the end of sequence symbol is generated.
% \textbf{Third}, we report the results for the temperature scaling method presented in \S \ref{subsec:uncertainty}. For comparison, we also compute the D-TP metric for the standard Transformer model on the subset of 400 segments considered for this experiment.
For comparison, we also compute the D-TP metric for the standard Transformer model on the subset of 400 segments considered for this experiment.

\input{tables/nmt_models_with_da.tex}

% Table \ref{tab:uncertainty_across_models} shows the results. Interestingly, correlation between output probabilities and DA is not necessarily related to the quality of MT outputs. For example, sampling produces much higher correlation although the quality is much lower. This is in line with previous work that indicates that sampling results in better calibrated probability distribution than beam search \cite{ott2018analyzing}. System variants that promote diversity in NMT outputs (diverse beam search and MoE) do not achieve any improvement in correlation over standard Transformer model, whereas ensemble model achieves the best results both in terms of MT quality and in terms of QE. We also note that applying temperature scaling improves the correlation without any drop in quality.

Table \ref{tab:uncertainty_across_models} shows the results. Interestingly, the correlation between output probabilities and DA is not necessarily related to the quality of MT outputs. For example, sampling produces much higher correlation although the quality is much lower. This is in line with previous work that indicates that sampling results in better calibrated probability distribution than beam search \cite{ott2018analyzing}. System variants that promote diversity in NMT outputs (diverse beam search and MoE) do not achieve any improvement in correlation over standard Transformer model.

The best results both in quality and QE are achieved by ensembling, which provides additional evidence that better uncertainty quantification in NMT improves correlation with human judgments. MC dropout achieves very similar results. We recommend using either of these two methods for %building
NMT systems with unsupervised QE.

\subsection{NMT Calibration across Training Epochs}

\begin{figure}[hbt!]
\centering
\includegraphics[scale=0.4]{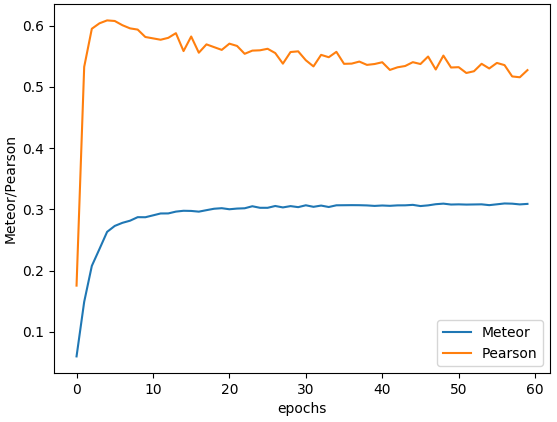}
\caption{Pearson correlation between translation quality and model probabilities (orange), and  Meteor (blue) over training epochs.}
\label{fig:calibration_by_epoch}
\end{figure}

The final question we address is how the correlation between translation probabilities and translation quality is affected by the amount of training. We train our base Et-En Transformer system for 60 epochs. We generate and evaluate translations after each epoch. We use the test partition of the MT training set and assess translation quality with Meteor evaluation metric. Figure \ref{fig:calibration_by_epoch} shows the average Meteor scores (blue) and Pearson correlation (orange) between segment-level Meteor scores and translation probabilities from the MT system for each epoch.

Interestingly, as the training continues test quality stabilizes whereas the relation between model probabilities and translation quality is deteriorated. During training, after the model is able to correctly classify most of the training examples, the loss can be further minimized by increasing the confidence of predictions \cite{guo2017calibration}. Thus longer training does not affect output quality but damages calibration.

%% previous version:
% In addition, we study how the correlation with translation quality is affected by the number of training epochs. Longer training that leads to an increase in the accuracy of model predictions does not necessarily imply an increase in the calibration of its predictive probabilities or may even damage it \cite{guo2017calibration}.%\todo{LS: this is another good candidate to go}

% It is not feasible to collect human judgments for translations produced by the system at each training epoch. Therefore, we generate translations for the test partition of the training set of the Estonian-English system and compute segment-level Meteor scores for the translations generated by our base Transformer system for 60 training epochs. We compute the average Meteor scores and the correlation between segment-level Meteor scores and segment-level translation probabilities. Figure \ref{fig:calibration_by_epoch} shows the results. 

% Recall from Section \ref{subsec:data} that our training set for Estonian-English is relatively small with only 900K sentence pairs. Training a system for many epochs can induce overfitting. Interestingly, as the training continues test quality stabilizes whereas the relation between model probabilities and translation quality is deteriorated. During training after the model is able to correctly classify most of the training examples, the loss can be further minimized by increasing the confidence of predictions \cite{guo2017calibration}.

%% file: tables/nmt_models_with_da.tex
\begin{table}[ht]
\begin{center}
%\small
\footnotesize
\begin{tabular}{l  c c } 
\toprule 
{\bf Method} & {\bf $r$ }& {\bf DA}\\ 
\hline
{TP-Beam} & 0.482 & 58.88 \\
\hline
{TP-Sampling} & 0.533 & 42.02 \\
{TP-Diverse beam} & 0.424 & 55.12 \\
\hline
{TP-RNN} & 0.502 & 43.63 \\
{TP-Ensemble} & 0.538 & 61.19\\
% {TP-MoE }& 0.429 & 51.20 \\
{TP-MoE} & 0.449 & 51.20 \\
\midrule
% TP-Temp & 0.536 & 58.23 \\
{D-TP} & 0.526 & 58.88 \\
% TP-Multilingual & 0.428 & 72.64 \\ % showing absolute correlation
\bottomrule
\end{tabular}
\end{center}
\caption{Pearson correlation ($r$) between sequence-level output probabilities (TP) and average DA for translations generated by different NMT systems.}
\label{tab:uncertainty_across_models}
\end{table}

% \begin{table}[h]
% \begin{center}
% \small
% \begin{tabular}{|l | c c c |} 
% \hline
% & $r$ & $r_s$ & DA\\ 
% \hline
% Beam & 0.482 & 0.481 & 58.88 \\
% Sampling & 0.533 & 0.536 & 42.02 \\
% Diverse beam & 0.424 & 0.457 & 55.12 \\
% Temperature & 0.536 & 0.534 & 58.23 \\
% \hline
% RNN & 0.502 & 0.508 & 43.63 \\
% Ensemble & 0.538 & 0.528 & 61.19\\
% MoE & 0.429 & 0.426 & 51.20 \\
% MoE-all & 0.449 & 0.449 & 51.20 \\
% Multilingual & 0.428 & 0.426 & 72.64 \\ % showing absolute correlation
% \hline
% \end{tabular}
% \end{center}
% \caption{Pearson and Spearman correlation between softmax output probabilities and DA judgments for translations generated by different NMT models.}
% \label{tab:uncertainty_across_models}
% \end{table}

%% file: conclusions.tex
\section{Conclusions}

% QE is crucial for NMT, given the increasing use of these systems in a wide variety of contexts. This poses new challenges for QE, as it needs to provide accurate and fast quality estimates for texts in all kinds of genres, styles, topics, etc. QE has been typically addressed as a supervised learning task with large resource-heavy systems providing highly accurate results. However, such systems are limited by access to annotated in-domain data.% for supervised training.

%We have proposed to view QE task from a different perspective. 
We have devised an unsupervised approach to QE where no training or access to any additional resources besides the MT system is required. Besides exploiting softmax output probability distribution and the entropy of attention weights from the NMT model, we leverage uncertainty quantification for unsupervised QE. We show that, if carefully designed, the indicators extracted from the NMT system constitute a rich source of information, competitive with supervised QE methods.

We analyzed how different MT architectures and training settings affect the relation between predictive probabilities and translation quality. We showed that improved translation quality does not necessarily imply a stronger correlation between translation quality and predictive probabilities. Model ensemble have been shown to achieve optimal results both in terms of translation quality and when using output probabilities as an unsupervised quality indicator.

Finally, we created a new multilingual dataset for QE covering various scenarios for MT development including low- and high-resource language pairs. Both the dataset and the MT models needed to reproduce the results of our experiments are available at \url{https://github.com/facebookresearch/mlqe}. 

This work can be extended in many directions. First, our sentence-level unsupervised metrics could be adapted for QE at other levels (word, phrase and document). Second, the proposed metrics can be combined as features in supervised QE approaches. Finally, other methods for uncertainty quantification, as well as other types of uncertainty, can be explored. % to achieve better performance.

%We note that while the paper covers QE at sentence level, the extension of our unsupervised metrics to QE at other levels (word, phrase, document) would be straightforward.

% By contrast to the few recent works that discuss calibration in NMT, we have analysed the relation between system predictive probabilities and human judgments of MT quality.

% We study the calibration of neural MT probability estimates. We introduce a variety of unsupervised quality measurements. By contrast to the few existing works on the topic, we use human judgments to evaluate the accuracy of model confidence estimates.

% We have approached QE from uncertainty quantification perspective. We have proposed various ways of quantifying uncertainty based on the softmax output, with softmax output entropy giving the best results. We further leveraged methods from approximate Bayesian inference showing that predictive probabilities obtained through model ensembling and MC dropout achieve the best results in predicting translation quality.